\documentclass[10pt,twocolumn,letterpaper]{article}

\usepackage{cvpr}              

\usepackage{amssymb}
\usepackage{pifont}
\newcommand{\cmark}{\ding{51}}%
\newcommand{\xmark}{\ding{55}}%
\usepackage{xcolor, colortbl}
\usepackage{times}
\usepackage{epsfig}
\usepackage{graphicx}
\usepackage{amsmath}
\usepackage{amssymb}
\usepackage{bm}
\usepackage{booktabs}
\usepackage{setspace}
\usepackage{float}
\usepackage{color}
\usepackage{multirow}

\definecolor{cyan}{cmyk}{.3,0,0,0}
\usepackage[pagebackref,breaklinks,colorlinks]{hyperref}

\usepackage[capitalize]{cleveref}
\crefname{section}{Sec.}{Secs.}
\Crefname{section}{Section}{Sections}
\Crefname{table}{Table}{Tables}
\crefname{table}{Tab.}{Tabs.}

\usepackage{soul}
	\definecolor{airforceblue}{rgb}{0.36, 0.54, 0.66}

\def\cvprPaperID{} 
\def\confName{CVPR}
\def\confYear{2022}

\begin{document}
	
	\title{End-to-End Human-Gaze-Target Detection with Transformers}
	
	\author{Danyang Tu$^1$, 
	        Xiongkuo Min$^1$, 
	        Huiyu Duan$^1$, 
	        Guodong Guo$^2$, 
	        Guangtao Zhai$^{1}$,
	        Wei Shen$^3$ \\
		$^1$Institute of Image Communication and Network Engineering, Shanghai Jiao Tong University\\
        $^2$Institute of Deep Learning, Baidu Research, Beijing, China \\
		$^3$MoE Key Lab of Artificial Intelligence, AI Institute, Shanghai Jiao Tong University\\
		{\tt\small \{danyangtu, minxiongkuo, huiyuduan, zhaiguangtao, wei.shen\}@sjtu.edu.cn, guoguodong01@baidu.com} \\
	}
	\maketitle
	\begin{abstract}
	    In this paper, we propose an effective and efficient method for Human-Gaze-Target (HGT) detection, i.e., gaze following. Current approaches decouple the HGT detection task into separate branches of salient object detection and human gaze prediction, employing a two-stage framework where human head locations must first be detected and then be fed into the next gaze target prediction sub-network. In contrast, we redefine the HGT detection task as detecting human head locations and their gaze targets, simultaneously. By this way, our method, named Human-Gaze-Target detection TRansformer or HGTTR, streamlines the HGT detection pipeline by eliminating all other additional components. HGTTR reasons about the relations of salient objects and human gaze from the global image context. Moreover, unlike existing two-stage methods that require human head locations as input and can predict only one human's gaze target at a time, HGTTR can directly predict the locations of all people and their gaze targets at one time in an end-to-end manner. The effectiveness and robustness of our proposed method are verified with extensive experiments on the two standard benchmark datasets, GazeFollowing and VideoAttentionTarget. Without bells and whistles, HGTTR outperforms existing state-of-the-art methods by large margins (6.4 mAP gain on GazeFollowing and 10.3 mAP gain on VideoAttentionTarget) with a much simpler architecture.
	\end{abstract}
	
	\vspace{-0.8cm}
	\section{Introduction}
	\label{sec:intro}
	Gaze following plays a crucial role in high level human-scene understanding tasks, and has attracted considerable research interest recently. Given an image or video frame containing one or more humans, the goal of gaze following is to predict where each person is looking at.
	
	\begin{figure}[t]
		\centering
		\includegraphics[width=\linewidth]{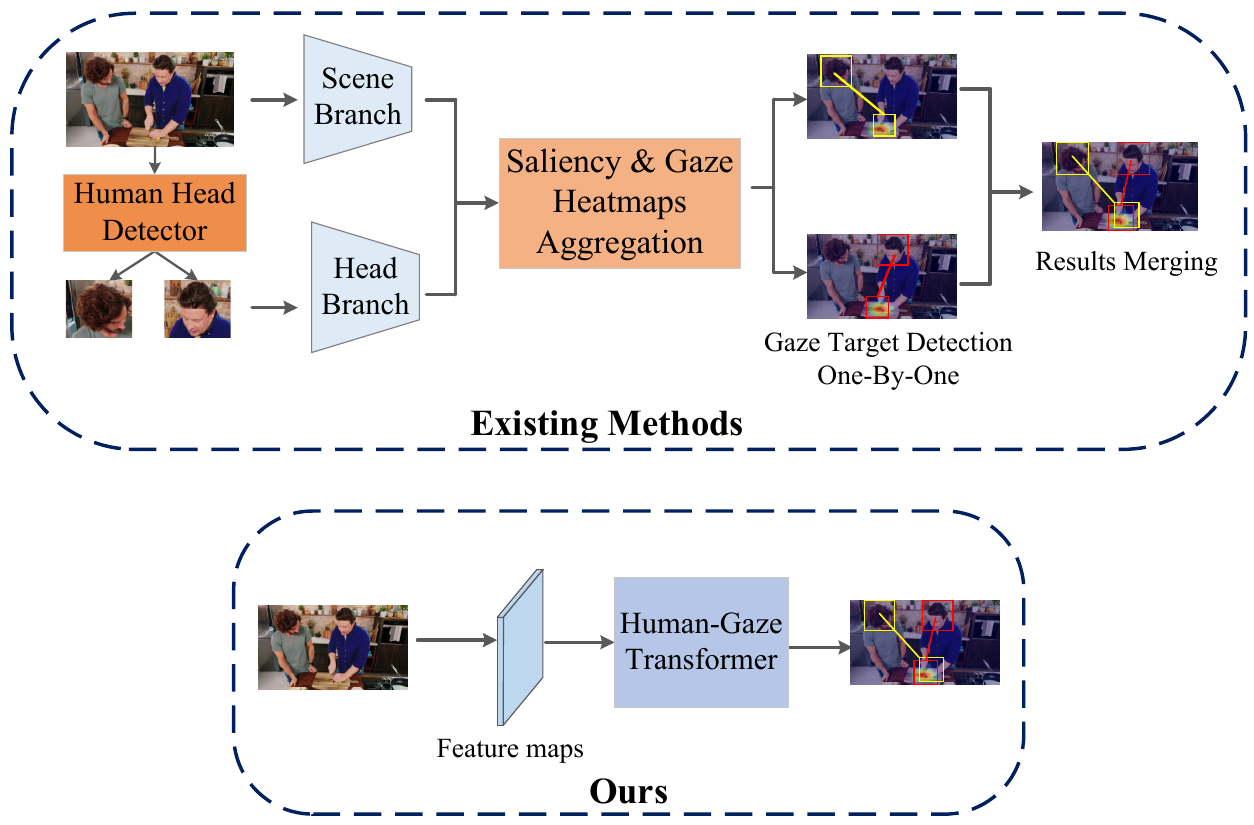}
		\caption{{\bfseries Pipelines of existing methods and our proposed model}.} 
		\vspace{-0.6cm}
		\label{first_img}
	\end{figure}
	
	Unlike traditional gaze prediction tasks~\cite{kr,huang2017tabletgaze} that predict only the gaze direction with a cropped human head image as input, gaze following further predicts the specific location in the scene that human is looking at. To this end, recent works leverage head pose features and the saliency maps of potential gaze targets by taking both head crops and the scene image as inputs. For instance, Recasens \etal~\cite{recasens} precisely detected the attention target of each person by extracting features from the scene and head images simultaneously. More recently, Chong \etal \cite{chong} proposed a novel framework to solve the problem of identifying gaze targets in videos. There are also other related works, including but not limited to \cite{chong2018, judd, leifman, zhu, kr, zhao2019, fang2021dual}. These methods are very attractive since they demonstrate the ability to estimate gaze targets directly from images or videos, without the help of any  monitor-based and wearable eye tracker devices.
	
	However, as shown in Figure~\ref{first_img}, existing methods share a similar multi-stream architecture, which contains a scene branch for scene understanding and another parallel head branch to extract head pose features. In this case, a common problem arises in that both head images and scene images must be taken as inputs simultaneously. As a consequence, existing methods face the following three major drawbacks: (1) An additional human head detector is then essential in practical applications, which compels the entire framework to be two-stage, and the precision of the additional detector can seriously affect the final results of gaze target detection. (2) As head crops are required for the second stage, existing methods can only predict gaze targets sequentially, which is less efficient when there are multiple persons in the same scene. It implies that the detection process will be conducted repeatedly and it is necessary to perform some post-operations to merge the detected gaze targets of different subjects in the same scene. (3) Most importantly, even if both head crops and scene images are taken as inputs, existing methods predict saliency maps and gaze directions separately, lacking contextual relational reasoning for interactions between them. 
	
	To  overcome these drawbacks, we propose HGTTR, a Human-Gaze-Target detection TRansformer that simultaneously detects a human's head location and his/her attention target by image-wide contextual modeling. More specifically, taking as input the image of a scene containing one or more humans, our HGTTR is designed to simultaneously detect the head locations of all individuals as well as their gaze targets at one time. Its outputs can be represented as $N$ human-gaze-target (HGT) instances in the format of $\langle${\it \text{head location, attention target}}$\rangle$, where $N$ is the number of persons in the image. Besides, with Transformers~\cite{attention} as key component,  HGTTR has a great ability to reason long-range gaze behaviors, thanks to its global contextual modeling capability.
	
	\begin{figure*}[!t]
	\vspace{-0.4cm}
		\centering
		\includegraphics[width=0.97\linewidth]{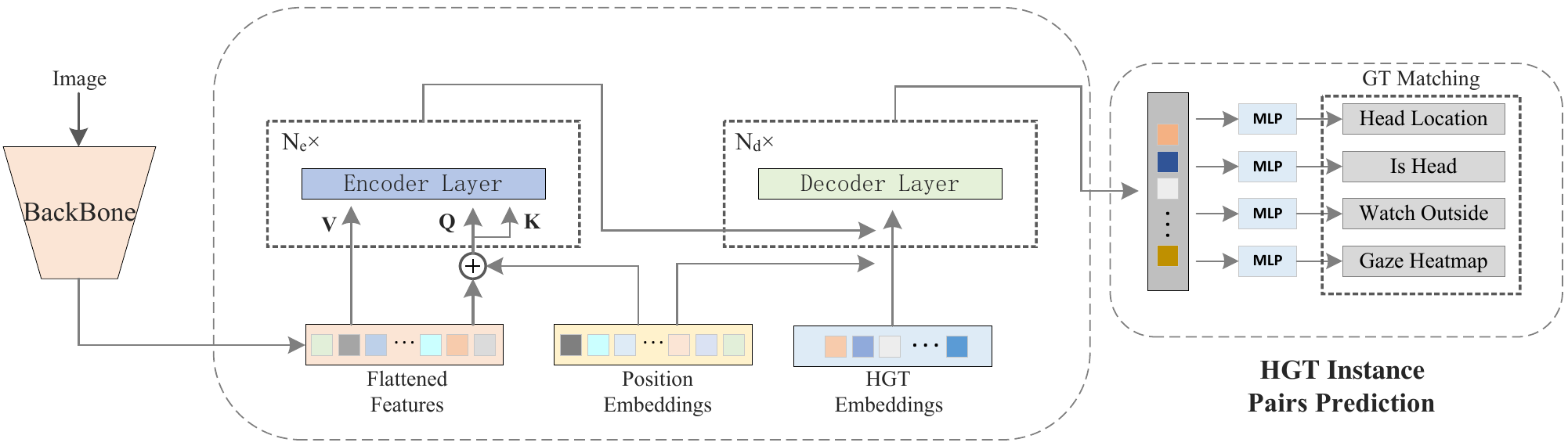}
		\caption{{\bfseries Pipeline of the proposed model}. It consists of four key components: a backbone, a typical Transformer, four multi-layer perceptions (MLP) and a quaternate loss function.} 
		\label{whole_view}
		\vspace{-0.4cm}
	\end{figure*}
	
	To this end, we reformulate HGT detection as a set-based prediction problem. We define a HGT query set with several learnable embeddings, and each query is designed to capture at most one HGT instance. The HGTTR first takes a CNN backbone to extract high-level image features from only a single scene image, and then the encoder is leveraged to generate global memory features by modeling the relation between the image features explicitly. After that, the HGT queries and the global memory features are sent to the decoder to generate the output embeddings. Finally, the HGT instances are predicted based on the output embeddings of the decoder with a multi-layer perception. Meanwhile, we also propose a quaternate HGT matching loss to supervise the learning process of HGT instances prediction. Experimental results show that HGTTR outperforms existing state-of-the-art methods by large margins. Specifically, it achieves a 6.4 mAP gain on GazeFollowing~\cite{recasens} and a 10.3 mAP gain on VideoAttentionTarget~\cite{chong} with a much higher FPS (more than 5 times compared to existing methods).
	
	\vspace{-0.3cm}

	\section{Related Work}
	\noindent
	\noindent
	{\bfseries Gaze following.} Gaze following was first proposed in \cite{recasens}, which presented a large dataset, GazeFollowing, and an algorithm accordingly. Unlike eye tracking~\cite{zhu, kr, zhu2017, zhang} and saliency detection~\cite{leifman, judd}, the goal of gaze following detection is to estimate what is being looked at by a person in a picture or a video frame. Based on this, Chong {\em et al.} \cite{chong2018} further addressed the problem of out-of-frame gaze targets by learning gaze angle and saliency simultaneously. By utilizing other auxiliary information, such as body pose \cite{guan2020}, sight lines \cite{zhao2019, lian2018}, the within-frame gaze target estimation can be further enhanced. Besides, the work of \cite{recasens2017} inferred gaze targets from videos. More recently, Chong {\em et al.} \cite{chong} proposed another new dataset named VideoAttentionTarget, which modeled the dynamics of gaze from video data and inferred per-frame gaze targets.
	In~\cite{fang2021dual}, a three-stage method was proposed to simulate the human gaze inference behavior in 3D space. 
	
	However, in both GazeFollowing and VideoAttentionTarget, human head locations were carefully and manually annotated and all existing methods took them as inputs, which is impossible for real world applications.
	
	\noindent
	{\bfseries Transformer.} Transformer was first proposed in Natural Language Processing (NLP) domain~\cite{attention}. With self-attention mechanism as key component, it has great ability to selectively capture long range dependence among all tokens. Recently, a great number of excellent works have been proposed that perform Transformer structure in vision tasks. In DETR~\cite{carion2020end}, object detection was reformulated as a set prediction problem and solved via a typical Transformer, which eliminates the need for many hand-designed components in object detection while demonstrating good performance. ViT~\cite{dosovitskiy2020image} solved the image classification by representing  an image as $16 \times 16$ patches and utilizing a Transformer encoder to predict the possible category. Transformer has also show great potential in many other vision tasks, such as semantic segmentation~\cite{zheng2021rethinking, strudel2021segmenter}, low-level vision tasks~\cite{chen2021pre} and so on.
	\vspace{-0.3cm}
	\section{Method}
	\subsection{Problem Reformulation}
	Given a single image $\bm{x} \in \mathbb{R} ^{3 \times H \times W}$ that contains one or more humans as input, we aim to predict all positions of all humans at one time, as well as their gaze locations. For convenience, we denote the position of people $\bm{l}_h$ as their head locations since they have been annotated in existing datasets. Specifically, a human's head location is represented as a bounding box $(x_{tl},y_{tl},x_{br},y_{br})$, where the subscripts $tl$ and $br$ denote top-left and bottom-right, respectively. Meanwhile, the gaze location $\bm{l}_g$ is represented as a Gaussian heatmap. By this means, our problem can be formulated as maximizing a joint \emph{posteriori} of the output pairs $\langle \bm{l}_h, \bm{l}_g \rangle$ on the input image:
	\begin{equation}\label{pro}
	\setlength{\abovedisplayskip}{3pt}
    \setlength{\belowdisplayskip}{3pt}
	\mathcal{T}^* \doteq \underset{\mathcal{T}}{\max}\,\prod_{i=1}^{N}p(\langle \bm{l}_h, \bm{l}_g \rangle_i\mid\bm{x}),
	\end{equation}
	where $N$ is the number of the humans in this image and $\mathcal{T}^*$ refers to the optimal model.
	
	It has to be noticed that this is quite different from the traditional gaze following problem, which takes both scene image and cropped human head image as input, and predicts only one individual's gaze location at a time. It can be formulated as:
	\begin{equation}\label{gazefollow}
	\setlength{\abovedisplayskip}{3pt}
    \setlength{\belowdisplayskip}{3pt}
	\mathcal{T}^* \doteq \underset{\mathcal{T}}{\max}\,p(\bm{l}_g^j \mid\bm{x},\bm{x}_h^j),
	\end{equation}
	where  $\bm{x}_h^j$ refers to the $j$-th cropped human head image and $\bm{l}_g^j$ denotes the $j$-th human's gaze location. Namely, for existing methods, an identical scene image is regarded as $N$ different cases when there are $N$ different humans.
	\vspace{-0.3cm}
	\subsection{Network Architecture}
	Figure~\ref{whole_view} illustrates the overall architecture of the proposed HGTTR. It consists of four main parts: (i) a backbone to extract high-level visual feature from the input image, (ii) a Transformer encoder-decoder to digest visual feature and generate output embeddings, (iii) several multi-layer perceptions (MLP) to predict HGT instances and (iiii) a quaternate loss function for bipartite matching.
	
	\noindent
	{\bfseries Backbone.} Given an input image $\bm{x} \in \mathbb{R}^{3\times H \times W}$, a feature map $\bm{z_b} \in \mathbb{R}^{D_b \times H' \times W'}$ is calculated by an arbitrary CNN backbone network. $\bm{z_b}$ is then fed into a projection convolution layer with a kernel size of $1 \times 1$ to reduce the dimension from $D_b$ to $D_c$. After that, a flatten operator is used to collapse the spatial dimension into one dimension, and a feature map $\bm{z}_f \in \mathbb{R}^{D_c \times HW}$ is obtained, which is denoted as \emph{Flattened Features} in Figure~\ref{whole_view}. In this work, we use ResNet~\cite{he2016} as our backbone and reduce the dimension of feature layer-5 from $D_b=2048$ to $D_c=256$.
	
	\noindent
	{\bfseries Encoder.} The encoder consists of $N_e$ encoder layers built upon standard Transformer structure with a multi-head self-attention (MSA) module and a feed-forward network (FFN). It takes the output of the backbone $\bm{z}_f \in \mathbb{R}^{D_c \times HW}$ as input to produce another feature map with richer contextual information. Besides, a fixed positional encoding $\bm{p} \in \mathbb{R}^{D_c \times H' \times W'}$ is additionally fed into the encoder to supplement the positional information as the Transformer architecture is permutation invariant. The attention in Transformer can be formulated as:
	\begin{equation}
	\setlength{\abovedisplayskip}{3pt}
    \setlength{\belowdisplayskip}{3pt}
	\begin{split}
	\label{eq:att}
	\mathrm{Attention}(Q, K, V) = \mathrm{softmax}(\frac{QK^T}{\sqrt{d_k}})V,\\
	Q=Q_f+Q_p, K=K_f+K_p, V=V_f,
	\end{split}
	\end{equation}
	where $d_k$ is the channel dimension, subscript $f$ means the feature and $p$ refers to the position encoding. The $Q, K, V$ are the query, key and value, respectively. As shown in Figure~\ref{whole_view}, in the self-attention module of the encoder, $Q_f = K_f = V_f = \bm{z}_f$, and $Q_p = K_p = \bm{p}$.
	
	\noindent
	{\bfseries Decoder.} We also build the decoder layer on the basis of transformer architecture. Different from the encoder layer containing only self-attention module, the decoder layer consists of both self-attention and cross-attention mechanisms. In self-attention, $K_f$ and $V_f$ are the same as $Q_f$ while $K_p$ is the same as $Q_p$. Specifically, $Q_f \in \mathbb{R}^{N_q \times C}$ is the output of the last decoder and we initialize it for the first decoder with a constant vector. For HGT query position embedding $Q_p \in \mathbb{R}^{N_q \times C}$, we use a set of learning embedding:
	\begin{equation}\label{query}
	\setlength{\abovedisplayskip}{3pt}
    \setlength{\belowdisplayskip}{3pt}
	Q_p = \text{Embedding}(N_q, C),
	\end{equation}
	where $N_q$ is always larger than the number of actual HGT instances in an image. In cross-attention, $Q_f \in \mathbb{R}^{N_q \times C}$ is generated from the output of the former self-attention module while $K_f \in \mathbb{R}^{HW \times C}$ and $V_f \in \mathbb{R}^{HW \times C}$ are the output features of the encoder. $Q_p$ is also set as Eq. (\ref{query}) and $K_p$ is equal to $\bm{p}$.
	
	In summary, the decoder has three inputs, the global memory from the encoder, HGT instance queries and position encoding. It serves to transform $N_q$ learnable position embeddings (denoted as \emph{HGT instance Embeddings} in Figure~\ref{whole_view}) into $N_q$ output embeddings with both the self-attention and cross-attention mechanisms. 
	
	\noindent
	{\bfseries MLP for HGT instances prediction.} Mathematically, we define each HGT instance by the following three vectors: a human-head-bounding-box vector normalized by the corresponding image size $\bm{l}_h \in [0,1]^4$, a watch-in-out (whether the gaze target is located inside the scene image or not) binary one-hot vector $\bm{w} \in \{0,1\}^2$, and a gaze heatmap vector $\bm{l}_g \in [0,1]^{H_o \times W_o}$, where $H_o$ and $W_o$ denote the spatial resolution of the output gaze heatmap. 
	
	The output embedding for each HGT query is decoded to one HGT instance by several multi-layer perception (MLP) branches. Specifically, we use two one-layer MLP branches $f_h$ and $f_o$ to predict the human confidence (whether the predicted bounding box is a human or not) and watch-in-out confidence, respectively. Meanwhile, a three-layer MLP branch $f_{lh}$ is set to predict human-head-bounding-box as well as a five-layer MLP branch $f_{lg}$ to predict the gaze heatmap. We use a softmax function for all one-layer branches and a sigmoid function for box and heatmap prediction.
	\vspace{-0.2cm}
	\subsection{Loss Calculation}
	The loss calculation is composed of two stages: the bipartite matching between ground-truths and output predictions of the proposed network, and the loss calculation for the matched pairs.
	
	\noindent
	{\bfseries Bipartite matching.} We follow the training procedure of DETR~\cite{carion2020end} and use the Hungarian algorithm~\cite{kuhn1955hungarian} for the bipartite matching, which is designed to obviate the process of suppressing over-detection. First of all, we pad the ground-truth of HGT instances with $\varnothing $ (no instance) so that the size of ground-truths set becomes $N_q$.
	
	As illustrated in Figure~\ref{whole_view}, the model outputs a fixed-size set of $N_q$ HGT instance predictions, and we denote them as $O = o^i, i=1,2,\cdots,N_q$. Meanwhile, we use $T = t^i, i = 1,2,\cdots M, \varnothing_1, \varnothing_2,\cdots,\varnothing_{N_q-M}$ to represent the ground-truths, where $M$ is the real number of HGT instances in an image. Then, the matching process can be denoted as an injective function: $\omega_{T\rightarrow O}$, where $\omega(i)$ is the index of predicted HGT instance assigned to the $i$-th ground-truth. We define the matching cost as:
	\begin{equation}\label{match-cost}
	\setlength{\abovedisplayskip}{3pt}
    \setlength{\belowdisplayskip}{3pt}
	\mathcal{L}_{cost}=\sum_{i}^{N_q}\mathcal{L}_{match}(t^i,o^{\omega(i)}),
	\end{equation}
	where $\mathcal{L}_{match}(t^i,o^{\omega(i)})$ is a matching cost between the $i$-th ground-truth and the $\omega(i)$-th prediction.
	
	Specifically, the matching cost $\mathcal{L}_{match}(t^i,o^{\omega(i)})$ consists of four types of cost: the head-box-regression cost $\mathcal{L}_{box}$, is-head cost $\mathcal{L}_h$, watch-in-out cost $\mathcal{L}_w$ and gaze heatmap cost $\mathcal{L}_g$.
	$\mathcal{L}_{box}$ is box regression loss for human head box, and the weighted sum of GIoU~\cite{giou} loss and $L_1$ loss is used:
	\begin{equation}\label{box-cost}
	\setlength{\abovedisplayskip}{3pt}
    \setlength{\belowdisplayskip}{3pt}
	\mathcal{L}_{box} = \alpha_1 \left \| t_i^b - o_{\omega(i)}^b \right \| - \alpha_2 \text{GIoU}( t_i^b, o_{\omega(i)}^b),
	\end{equation}
	where the superscript $b$ refers to the bounding box. Besides, the $\mathcal{L}_h$ and $\mathcal{L}_w$ are respectively defined as:
	\begin{align}\label{is}
	\setlength{\abovedisplayskip}{3pt}
    \setlength{\belowdisplayskip}{3pt}
	&\mathcal{L}_h = -o_{\omega(i)}^c(k) \;\; s.t. \;\; t_i^c(k)=1,\\
	&\mathcal{L}_w = -o_{\omega(i)}^w(k) \;\; s.t. \;\; t_i^w(k)=1,
	\end{align}
	where $c \in \{0,1\}^2$ and $w \in \{0,1\}^2$ are one-hot vector for is-head and watch-in-out, respectively. We use $L_2$ loss for heatmap cost $\mathcal{L}_g$:
	\begin{equation}\label{gaze}
	\setlength{\abovedisplayskip}{3pt}
    \setlength{\belowdisplayskip}{3pt}
	\mathcal{L}_{g} =  \left \| t_i^g - o_{\omega(i)}^g \right \|_2.
	\end{equation}
	On this basis, we design the following matching cost for HGT prediction:
	\begin{small}
	\begin{equation}\label{match}
	\setlength{\abovedisplayskip}{3pt}
    \setlength{\belowdisplayskip}{3pt}
	\mathcal{L}_{match}(t^i,o^{\omega(i)}) = \beta_1 \mathcal{L}_{box} + \beta_2 \mathcal{L}_h + \beta_3 \mathcal{L}_w + \beta_4 \mathcal{L}_{g}.
	\end{equation}
	\end{small}
	
	We then leverage the Hungarian algorithm to determine the optimal assignment $\hat{\omega}$ among the set of all possible permutations of $N_q$ elements $\bm{\Omega}_{N_q}$. It can be formulated as:
	\begin{equation}\label{hun}
	\setlength{\abovedisplayskip}{3pt}
    \setlength{\belowdisplayskip}{3pt}
	\hat{\omega} = \underset{\omega \in \bm{\Omega}_{N_q}}{\arg \min}\mathcal{L}_{cost}.
	\end{equation}
	
	\noindent
	{\bfseries Loss function.} After the optimal one-to-one matching between the ground-truths and the predictions is found, the loss to be minimized in the training phase is calculated as Eq. (\ref{match}). The hyper-parameters are set as same as they does in matching process.
	
	\vspace{-0.3cm}
	\section{Experiments}
	
	\begin{table*}[htb]
  \small
  \centering
  
  \resizebox{\textwidth}{0.13\textheight}{ 
  \begin{tabular}{lcccccccccccccccc}
    \toprule[1pt]

    \multirow{3}{*}[-0.58em]{Method} &  
    
     \multicolumn{7}{c}{GazeFollowing} & & &
     
     \multicolumn{7}{c}{VideoAttentionTarget} \\
     
      \cmidrule(lr){2-8}
      \cmidrule(lr){11-17}
    & \multicolumn{2}{c}{AUC$\uparrow$} & \multicolumn{2}{c}{Average Dist.$\downarrow$} & \multicolumn{2}{c}{Min Dist.$\downarrow$} & \multirow{2}{*}[-0.5em]{\textbf{mAP}$\uparrow$} & &
    & \multicolumn{2}{c}{AUC$\uparrow$} & \multicolumn{2}{c}{L2 Dist.$\downarrow$}      & \multicolumn{2}{c}{AP$\uparrow$}        & \multirow{2}{*}[-0.5em]{\textbf{mAP}$\uparrow$}  
    \\
    \cmidrule(lr){2-3}
    \cmidrule(lr){4-5}
    \cmidrule(lr){6-7}
    \cmidrule(lr){11-12}
    \cmidrule(lr){13-14}
    \cmidrule(lr){15-16}
   & Default & \textbf{Real} & Default & \textbf{Real} & Default & \textbf{Real} & & & 
   & Default & \textbf{Real} & Default & \textbf{Real} & Default & \textbf{Real} & \\
    \midrule[0.7pt]
                
              Random %
              & 0.504 & 0.391 & 0.484 & 0.533 & 0.391 & 0.487 & 0.104 & &
              & 0.505 & 0.247 & 0.458 & 0.592 & 0.621 & 0.349 & 0.091
              \\
       
              Center %
              & 0.633 & 0.446 & 0.313 & 0.495 & 0.230 & 0.371 & 0.117 & &
              & ---   & ---   & ---   & ---   & ---   & ---   & ---
              \\
              
        	  Fixed bias %
        	  & ---   & ---   & ---   & ---   & ---   & ---   & --- & &
        	  & 0.728 & 0.522 & 0.326 & 0.472 & 0.624 & 0.510 & 0.130
        	  \\
        	  
        	  Judd~\cite{judd} %
        	  & 0.711 & ---   & 0.337 & ---   & 0.250 & ---   & --- & &
        	  & ---   & ---   & ---   & ---   & ---   & ---   & ---
        	  \\
        	  
        	  GazeFollow~\cite{recasens} %
        	  & 0.878 &0.804  & 0.190 & 0.233 & 0.113 & 0.124 & 0.457 & &
        	  & ---   & ---   & ---   & ---   & ---   & ---   & ---
        	  \\
        	  
        	  Chong~\cite{chong2018} %
        	  & 0.896 & 0.807 & 0.187 & 0.207 & 0.112 & 0.120 & 0.449 & &
        	  & 0.830 & 0.791 & 0.193 & 0.214 & 0.705 & 0.651 & 0.374
        	  \\
        	  Zhao~\cite{zhao2019} %
        	  & ---   &---   & 0.147  &---    & 0.082 &---    & --- & &
        	  & ---   & ---  & ---    & ---   & ---   & ---   & ---
        	  \\
        	  
        	  Lian~\cite{lian2018} %
        	  & 0.906 & 0.881 & 0.145 & 0.153 & 0.081 & 0.087 & 0.469 & &
        	  & 0.837 & 0.784 & 0.165 & 0.172 & ---   &---    & 0.392
        	  \\
        	  
        	  VideoAttention~\cite{chong} %
        	  & 0.921 & 0.902 & 0.137 & 0.142 & 0.077 & 0.082 & 0.483 & &
        	  & 0.860 & 0.812 & 0.134 & 0.146 & 0.853 & 0.849 & 0.420
        	  \\
        	  
        	  DAM~\cite{fang2021dual} %
        	  & 0.922 & ---   & 0.124 & ---   & 0.067 & ---   & --- & &
        	  & 0.905 & ---   & 0.108 & ---   & 0.896 & ---   & ---
        	  \\
        	  
        \midrule
        \rowcolor{cyan!50} 
        	  HGTTR (ResNet-50)
        	  & ---  & \textbf{0.917}   & ---  & \textbf{0.133} & ---   & 0.069 & \textbf{0.547} & &
        	  & ---  & 0.893   & ---  & 0.137 & ---   & 0.821 & 0.514
        	  \\
        \rowcolor{cyan!50} 
        	  HGTTR (ResNet-101)
        	  & ---  & 0.905  & ---  & 0.138 & ---   & \textbf{0.065} & 0.541 & &
        	  & ---  & \textbf{0.904}   & ---  & \textbf{0.126} & ---   & \textbf{0.854} & \textbf{0.523}
        	  \\
        \bottomrule[1pt]

  \end{tabular}
  }
  \caption{
  	\textbf{Quantitative comparisons on the GazeFollowing and VideoAttentionTarget sets.} 
  	As the existing methods require to take as input both scene and head images, we report the results of them in `Default' and `Real' set, respectively. Specifically, `Default' refers to use the head location that carefully labeled in existing dataset, while `Real' represents to apply an additional head detection network to predict the head location for real world applications.
  }
\label{tab_all}
\vspace{-0.4cm}
\end{table*}

	\subsection{Datasets and Evaluation Metrics}
	
	\noindent
	{\bfseries Datasets.} We train and test our model on both GazeFollowing~\cite{recasens} dataset and VideoAttentionTarget~\cite{chong} dataset. Specifically, we use every single frame in VideoAttentionTarget dataset as input during the training process, without considering the temporal information. Therefore, to avoid overfitting, for every five continuous frames from the training set of VideoAttentionTarget which have no obvious appearance differences, we randomly select one for training since they have almost the same gaze target. For testing, we still use all images in the testing set.
	
	Moreover, one of objectives of our proposed model is to predict the locations of different individuals. Therefore, the head locations in existing dataset annotations are no longer used as inputs but as ground-truths. Besides, previous works predict the gaze target for different subjects in an identical scene on a case-by-case basis, which results in each image being assigned with $M$ annotations, and $M$ is the number of people in an image. Unlike that, we merge the annotations of the same image into the same format as COCO~\cite{lin2014microsoft} object detection since we aim to predict them all at one time.
	
	\noindent
	{\bfseries Evaluation metric.} In this work, we have to evaluate the performance of proposed model in terms of both gaze target detection and human position detection. 
	
	For the former, we follow the standard evaluation protocols, as in \cite{recasens,chong}, to report the results in terms of $\textbf{AUC}$ and $\bm{L_{2}}$ distance. AUC: The final heatmap provides the prediction confidence score which is evaluated at different thresholds in the ROC curve. The area under curve (AUC) of the ROC is reported \cite{chong}. Distance: $L_2$ distance between the annotated target location and the prediction given by the pixel having the maximum value in the heatmap, with image width and height normalized to 1. Specifically, since the ground truth for GazeFollowing may be multimodal, the $L_2$ distance is the Euclidean distance between our prediction and the average of ground-truth annotations. Besides, the minimum distance between our prediction and all ground-truth annotations is also reported. In addition, the average precision (\textbf{AP}) is used to evaluate the performance for is-watching-outside prediction.
	
	For the latter, we use the commonly used role \emph{mean average percision} (\textbf{mAP}) to examine the model performance on both datasets. Specifically, a HGT detection is considered as true positive if and only if it localizes the human and detects gaze target accurately (\emph{i.e.} the \emph{Interaction-over-Union} (IOU) ratio between the predicted human-head-box and ground-truth is greater than 0.5 while the $L_2$ distance for gaze target detection is less than 0.15).
	
	\subsection{Implementation Details}
	The experiments are conducted on two popular backbone: ResNet-50 and ResNet-101~\cite{he2016}. Both Transformer encoder and decoder consist of 6 layers with a multi-head self-attention of 8 heads. We initialize the network with the parameters of DETR trained with the COCO dataset. The model is trained for 150 epochs using AdamW~\cite{adamw} optimizer with batch size of 16. Specifically, the initial learning rate of the backbone network is set as $10^{-5}$ while that of the others is set as $10^{-4}$, the weight decay is equal to $10^{-4}$. For the hyper-parameters of the model, we set $\alpha_1$ and $\alpha_2$ as $1.0$ and $2.5$, respectively. The weights ($\beta_1 - \beta_4$) for different cost functions in the loss are set as 2, 1, 1, 2, respectively. The number of HGT instance queries $N_q$ is 20 for all datasets. Both learning rates are decayed after 80 epochs. All experiments are conducted on 8 NVIDIA GTX 2080TI GPU.
	
	\subsection{Comparison to State-of-the-Art}
	We first show the main quantitative comparison of our HGTTR with the latest HGT detection methods in Table \ref{tab_all}. Specifically, since the cropped head image and the head location are essential for existing methods, we report the results of them on two different settings: `Default' and `Real'. For `Default', we directly utilize the carefully labeled head locations in existing dataset annotations. Meanwhile, we employ an additional head detector to automatically generate the head positions and feed the predicted results into existing models for real world applications, and the results are reported as `Real'. Following \cite{chong}, we fine-tuned a SSD-based~\cite{liu2016ssd} head detection network with the head annotations in existing dataset. As can be seen from the table, HGTTR outperforms existing state-of-the-art methods on all datasets. HGTTR with the ResNet-50 backbone yields a significant gain of 6.4 mAP compared with VideoAttention~\cite{chong} and 7.8 mAP compared with GazeFollow~\cite{recasens} on the GazeFollowing datasets. Moreover, HGTTR performs better on the VideoAttentionTarget dataset. This can mainly attribute to two main reasons: 1) VideoAttentionTarget is a larger dataset than GazeFollowing, which is important for transformer-based methods. 2) There are more than one humans in each image of VideoAttentionTarget while every image in GazeFollowing contains only one person. As our model outputs a fixed number of HGT instances, fewer instances in an image imply a higher value of fault positive (FP). In terms of gaze target prediction, our HGTTR still has outstanding performance. For VideoAttentionTarget dataset, we achieve a gain of 8 AUC compared with VideoAttention for `Real' setting.
	
	\vspace{-0.2cm}
	\subsection{Ablation Study}
	\vspace{-3pt}
	In this section, we conduct extensive experiments to validate the effectiveness of our proposed HGTTR. The ablation experiments are conducted with ResNet-50 backbone model, and the results are reported on the VideoAttentionTarget dataset.
	
	\begin{figure}[t]
	\vspace{-0.5cm}
		\centering
		\includegraphics[width=0.98\linewidth]{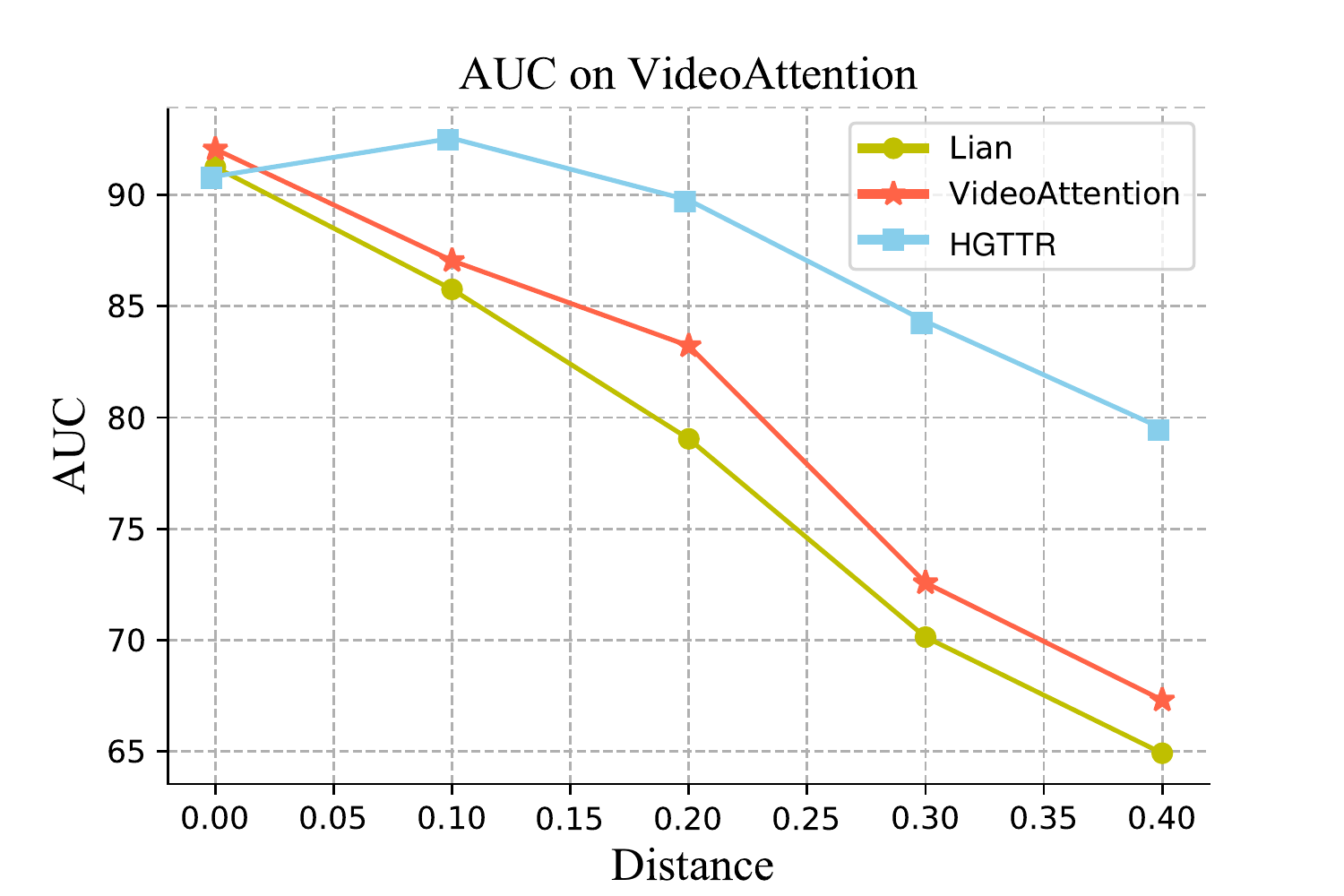}
		\caption{ {\bfseries Performance of different methods on different spatial distributions of HGT instances on VideoAttentionTarget.}}
		\vspace{-11pt}
		\label{dist}
	\end{figure}
    
    \vspace{-4pt}
    
	\vspace{-15pt}
	\subsubsection{Model components}
	\vspace{-3pt}
	\noindent
	{\bfseries Self-attention.} As the most important component in transformer, self-attention has great ability to capture long range contextual information. In \cite{chong}, attention mechanism has been simply applied to merge the features of head branch and scene branch, which achieved attractive performance. We first investigate in which cases self-attention especially achieves superior performance compared with the existing methods. To this end, we split HGT instances into bins of size 0.1 on the basis of $L_1$ distance between the center of a human's head and his gaze target, where the height and width of image have been normalized. The AUC of different methods in each bin is shown in Figure \ref{dist}, where the results of Lian~\cite{lian2018} and VideoAttention~\cite{chong} are reported in `Default' setting. The relative gaps of the performance between HGTTR and the other two methods become more evident as the distance grows. It indicts that gaze target detection tends to become more difficult as the distance grows. Besides, it is especially difficult for existing methods to deal with the distant cases while HGTTR has relatively better performance. The possible reason for such result is that existing methods rely on limited receptive fields for the feature aggregation. They are weak in capturing long range contextual information or easily be dominated by irrelevant information in the distant cases. On the contrary, the features of HGTTR are more effective thanks to the ability of self-attention to adaptively extract image-wide contextual information.

\begin{figure}[t]
	    \vspace{-0.5cm}
		\centering
		\includegraphics[width=0.98\linewidth]{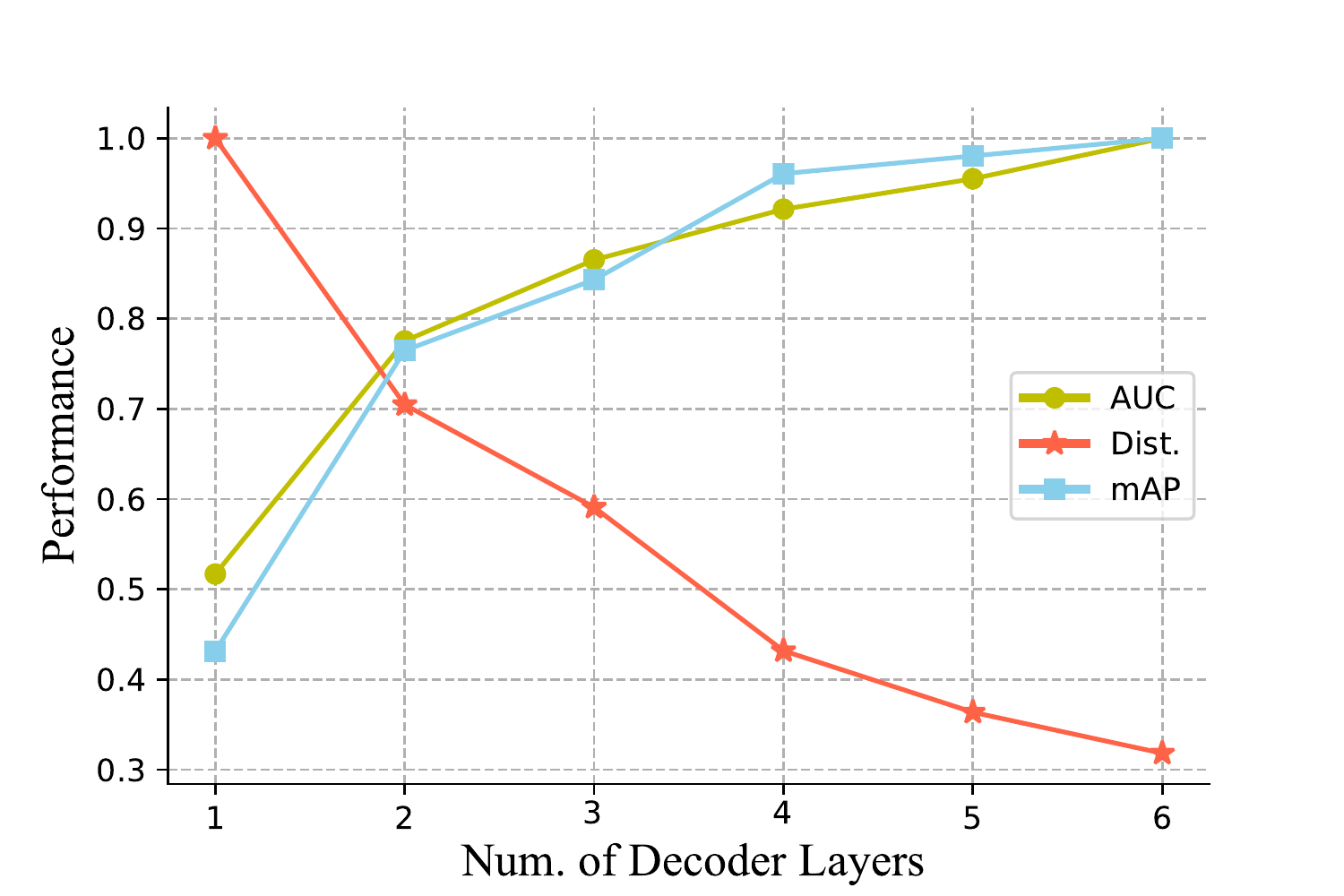}
		\caption{{\bfseries Performance of the model with different numbers of decoder layers.} To show them in the same figure, all metrics are normalized by dividing by the maximum value. Namely, we mainly show their trends as the number of decoder layers increases.}
		\label{decoder_num}
		\vspace{-0.3cm}
	\end{figure}

	\noindent
	{\bfseries Decoder.} A decoder is essential to transform the manually defined HGT queries $Q_f$ into a set of HGT instances with the features generated by the encoder. As each layer of the decoder is identical in the architecture, it is a coarse-to-fine process where each layer takes the predicted results of the previous layer as input to further produce more precise predictions. As shown in Figure~\ref{decoder_num}, HGTTR can achieve better performance with more decoder layers, especially for `AUC' and `$L_2$ Dist.'. A possible explain for this result is that the decoder can learn some potential relationships among different HGT instances with more self-attention mechanisms. For example, people in an identical scene more tend to have a similar gaze target.
	\begin{table}[t]
	\small
	\setlength{\tabcolsep}{8pt}
	\centering
	\begin{tabular}{cccc|ccc}
		\toprule[1pt]
		$\beta_1$   & $\beta_2$   & $\beta_3$   & $\beta_4$         & AUC$\uparrow$  & $L_2$ Dist.$\downarrow$  & mAP$\uparrow$  \\
		\midrule
		 1          & 1           & 1           & 1                 & 0.864          & 0.142                    & 0.492                \\
		 1          & 2           & 2           & 1                 & 0.857          & 0.148                    & 0.487                 \\
	\rowcolor{cyan!50}
		 2          & 1           & 1           & 2                 & \textbf{0.893}          & \textbf{0.137}                    & \textbf{0.514}                  \\
		 	
	\bottomrule[1pt]
		
	\end{tabular}
	\caption{{\bfseries The effects of different cost functions in matching process}. }
	\vspace{-0.7cm}
	\label{tab:matching}
\end{table}
	
	\begin{figure*}[t]
		\centering
		\includegraphics[width=0.97\linewidth,height=0.27\textheight]{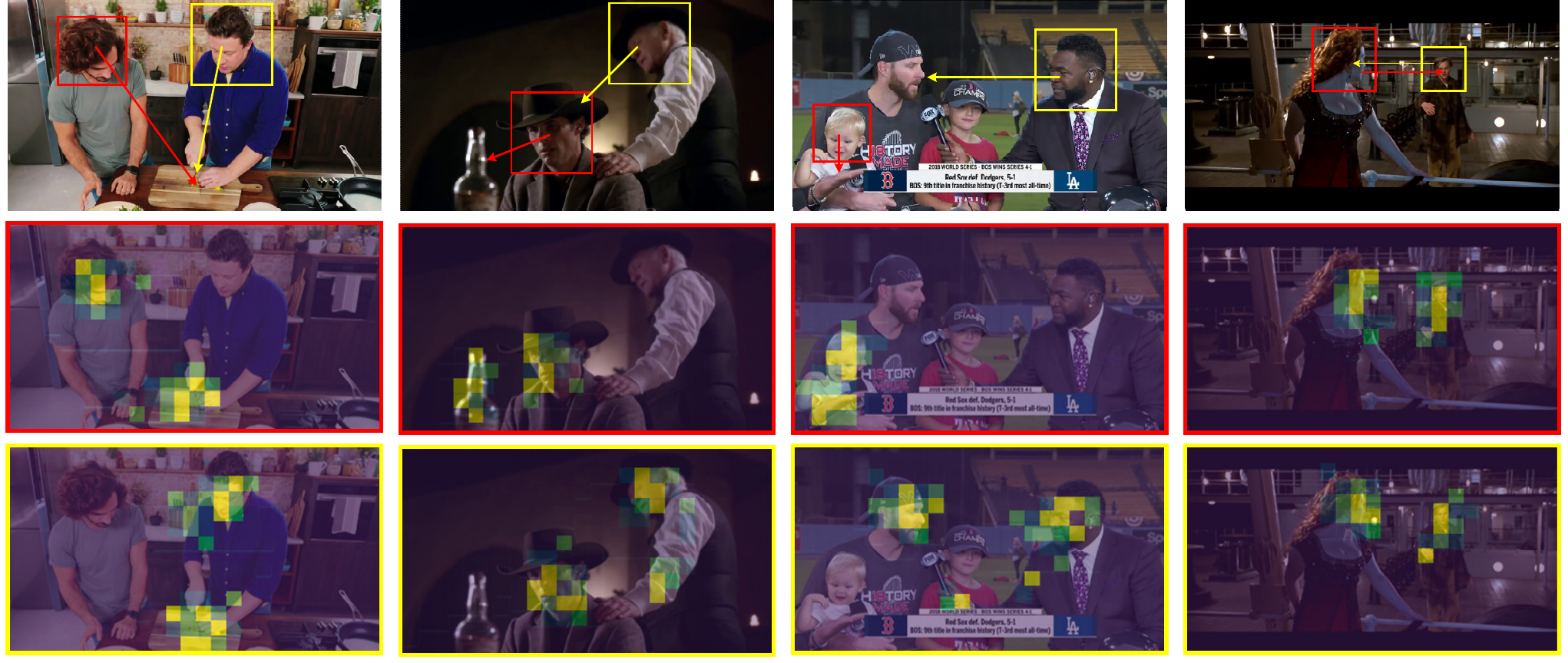}
		\caption{{\bfseries Visualization of attention maps in the decoder for the predicted HGT instances.} The images are randomly selected from the testing set of VideoAttentionTarget. As can be seen from the figure, our method has great ability to capture long range dependence. In addition, it can be seen from the figure that different HGT instances may share little common pattern (the last column), which indicts that an unique matching process is essential.}
		\label{attention_img}
	\end{figure*}

	\begin{table*}[t]
	\small
	\centering
	\resizebox{\linewidth}{0.09\textheight}{
	\begin{tabular}{l|c|c|p{0.01em}ccp{0.01em}ccp{0.01em}ccp{0.01em}cc}
		\toprule[1pt]
		\multicolumn{1}{l|}{\multirow{2}{*}[-0.30em]{Method}} & \multicolumn{1}{c|}{\multirow{2}{*}[-0.30em]{Fine-tuned}} & \multirow{2}{*}[-0.30em]{FPS} & & \multicolumn{2}{c}{Blur} & & \multicolumn{2}{c}{Gaussian Noise} & & \multicolumn{2}{c}{Brightness} & &\multicolumn{2}{c}{Normal}\\
		\cline{5-6} 
		\cline{8-9}
		\cline{11-12}
		\cline{14-15}
		& & & & AUC   & mAP   &   & AUC   & mAP   &   & AUC   & mAP   &   & AUC   & mAP \\
		\hline

		  \multicolumn{1}{c|}{\multirow{2}{*}{SSD-based~\cite{liu2016ssd}}}
		& \xmark   & 2   &   & 0.729   & 0.302   &   & 0.706   & 0.298   &   & 0.738   & 0.315   &   & 0.789   & 0.405 \\
		& \cmark   & 2   &   & 0.774   & 0.371   &   & 0.767   & 0.362   &   & 0.784   & 0.376   &   & 0.812   & 0.420  \\
		\hline
		\multirow{2}{*}{FCHD~\cite{vora2018fchd}}
		& \xmark   & 3   &   & 0.757   & 0.336   &   & 0.744   & 0.328   &   & 0.764   & 0.341   &   & 0.804   & 0.409 \\
		& \cmark   & 3   &   & 0.784   & 0.385   &   & 0.781   & 0.325   &   & 0.796   & 0.392   &   & 0.818   & 0.424  \\
		\hline
    		\multirow{2}{*}{HeadHunter~\cite{sundararaman2021tracking}}
		& \xmark   & 1   &   & 0.782   & 0.341   &   & 0.750   & 0.332   &   & 0.773   & 0.352   &   & 0.796   & 0.412 \\
		& \cmark   & 1   &   & 0.787   & 0.389   &   & 0.789   & 0.384   &   & 0.804   & 0.410   &   & 0.819   & 0.424  \\
		\hline
		\rowcolor{cyan!50}
		& \xmark   & 16  &   & 0.806   & 0.442   &   & 0.792   & 0.438   &   & 0.813   & 0.451   &     & 0.893   & 0.514 \\
		\rowcolor{cyan!50}
		\multirow{-2}{*}{HGTTR (Ours)}
		 & \cmark   & \textbf{16}  &   & \textbf{0.872}   & \textbf{0.487}   &   & \textbf{0.864}   & \textbf{0.481}   &   & \textbf{0.881}   & \textbf{0.496}   &     & ---     & ---    \\
		 	                                 
	\bottomrule[1pt]
		
	\end{tabular}
	}
	\caption{{\bfseries Performance of different human head detectors}. We manually degrade the original images in the testing set with several common distortion types. Specifically, `Fine-tuned' refers to that the training set is also augmented with these distortion types and the pre-trained head detector is further fine-tuned with the head locations in annotations. `Normal' denotes that no degeneration image is used.}
	\vspace{-0.5cm}
	\label{tab:head}
\end{table*}
	
	Moreover, we visualize the decoder attention map for the predicted HGT instances in Figure~\ref{attention_img} . The heatmap highlights both the human heads and their gaze targets, which indicts that our HGTTR reasons about the relations between human and scene from a more global image context.
	
	\noindent
	{\bfseries Matching strategy.} Our proposed matching cost function consists of four main aspects: human head location $\mathcal{L}_{box}$, whether a predicted bounding box is a human or not (Is Head) $\mathcal{L}_{h}$, whether the gaze target is located in the scene image or not (Watch Outside) $\mathcal{L}_{w}$ and the predicted gaze heatmap $\mathcal{L}_{g}$. Specifically, $\mathcal{L}_{box}$ and $\mathcal{L}_{g}$ are localization losses while $\mathcal{L}_{h}$ and $\mathcal{L}_{w}$ are classification losses. In this case, we conduct ablation study to further find the relative importance in matching: location first or classification first? In Eq. (\ref{match}), $\beta_1$ to $\beta_4$ denote the different weights for each matching cost function. As shown in Table~\ref{tab:matching}, the best result is obtained under $\beta_1 = \beta_4 = 2$ and $\beta_2 = \beta_3 = 1$, which suggests that localization plays a relatively more important role than classification during the matching process.

    \vspace{-7pt}
	\subsubsection{Importance of pairwise detection}
	\vspace{-3pt}
	We redefine the gaze following task to detect the human head locations and their gaze targets, simultaneously. In this subsection, we mainly analyze the necessity of this strategy.

	\noindent
	{\bfseries Model robustness.} As existing methods take both scene and human head images as inputs, an additional human head detector is essential for them. However, performance of gaze target prediction would then be seriously influenced by the precision of the head detector in this way. As shown in Table~\ref{tab:head}, we applied several different head detectors to analyze  model performance in different conditions. In real world applications, such as video surveillance, blur, noise, and brightness variation are the most common types of image degradation. It can be seen from the table, pairwise detection strategy is not only more efficient, but also more robust with better performance under different degradations. First, using an additional head detector is a sub-optimal solution since existing models are trained with head locations that are carefully and manually labeled so that they are very sensitive to the results of the head detector. Secondly, while noise can be addressed to some extent by data augmentation in real world applications, head detector is hard to fine-tune since it is impossible to manually generate annotations of head locations. On the contrary, we solve HGT detection in an absolute end-to-end manner, which does not require head location as input. Moreover, as shown in the table, HGTTR has better robustness with different image degradations. The possible reasons are that Transformer inherently has great ability to against noises and pairwise detection is not as sensitive to image degradations as two-stage methods.
	
	\begin{figure}[t]
	\vspace{-0.6cm}
		\centering
		\includegraphics[width=0.98\linewidth]{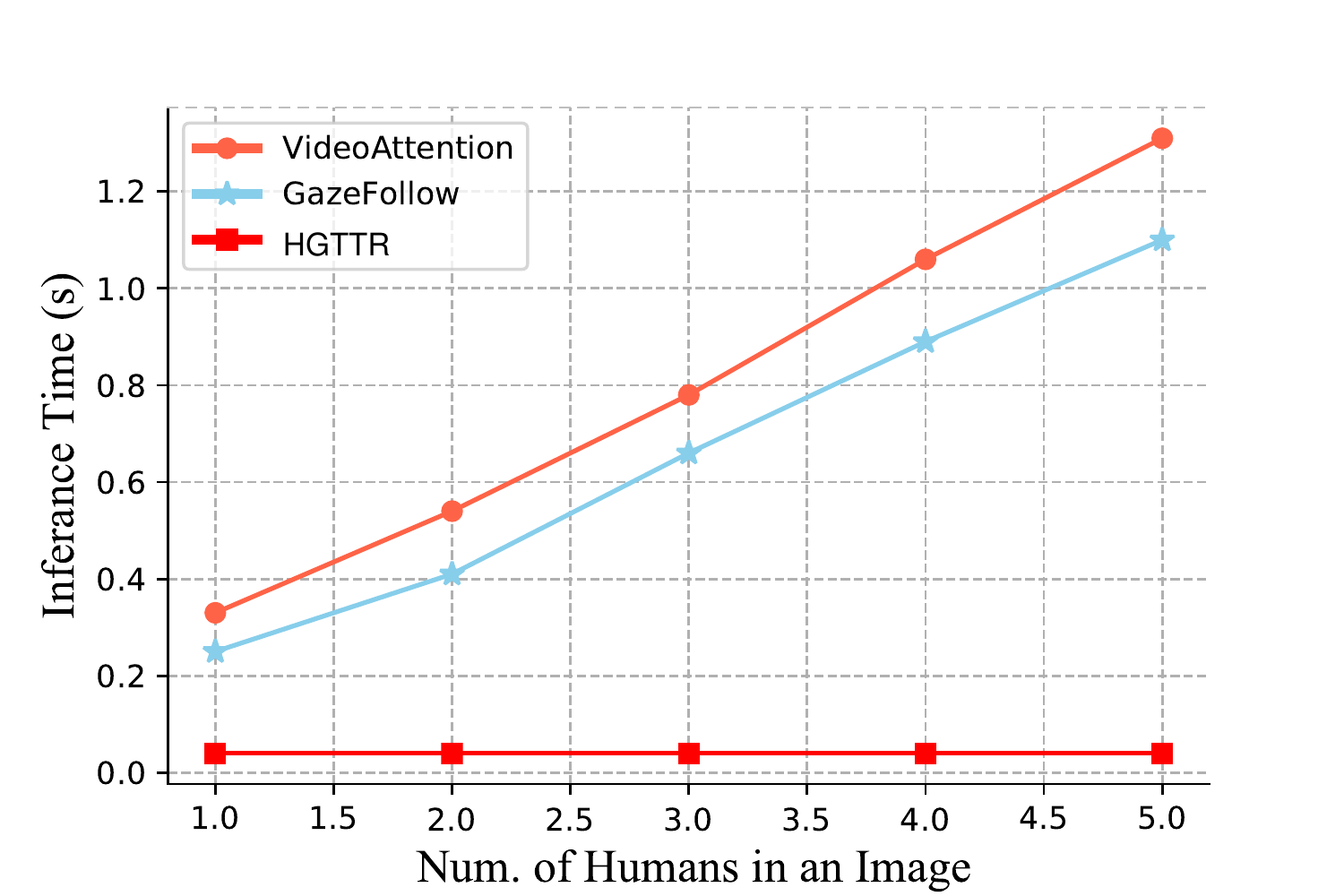}
		\caption{{\bfseries Inference time of different models with different numbers of humans in an image.}}
		\vspace{-0.4cm}
		\label{times}
	\end{figure}
	
	\noindent
	{\bfseries Model efficiency.} Another advantage of pairwise detection is high efficiency. Using human head images as input, as existing methods do, limits the ability of the model to predict different people's gaze targets at the same time. As shown in Figure~\ref{times}, as the number of individuals in an identical scene increases, the inference time of existing methods grows extremely. The main reason is that existing methods can predict only one human's gaze target at a time,  and the detection process has to be conducted repeatedly when there are more than one human in a same image. On the contrary, our proposed method has no such limitation and the inference time is almost not inflected by the number of humans. In this paper, our HGTTR is designed to detect up to 20 peoples gaze targets at a time. In real world applications, the model can be easily fine-tuned to increase the number of maximum detectable humans by simply changing the value of hyper-parameter $N_q$.
	\vspace{-0.2cm}
	\subsection{Qualitative Analysis}
	
	\noindent
	{\bfseries Practical application.} To evaluate our model's performance in practical applications, we randomly selected 15,000 pictures from the DL Gaze dataset~\cite{lian2018}, which records the daily activities of 16 volunteers in 4 different scenes, and the ground-truth is annotated by the observers in the videos. All models are only trained with the VideoAttentionTarget dataset. In addition, an additional human head detector is also employed for existing methods. The quantitative results are presented in Table~\ref{tab:prat}. Without bells and whistles, our HGTTR outperforms all state-of-the-art methods by a significant margin.
	
	\noindent
	{\bfseries Shared attention detection.} With the ability to detect gaze targets of different people at a same time, our methods are inherently suitable for inferring shared attention in social scenes. Following~\cite{chong}, the results are reported in terms of accuracy for interval detection of shared attention and $L_2$ distance for location prediction on the VideoCoAtt~\cite{fan2018inferring} dataset. This dataset has 113,810 test frames that are annotated with the target location when it is simultaneously attended by two or more people. As can be seen from Table~\ref{tab:share}, our model has demonstrated potential value of  recognizing higher-level social gaze.
	
	\begin{table}[t]
	\small
	\setlength{\tabcolsep}{6pt}
	\centering
	\begin{tabular}{l|cccc}
		\toprule[1pt]
		Methods         & AUC$\uparrow$  & $L_2$ Dist.$\downarrow$  & Ang.$\downarrow$ & mAP$\uparrow$  \\
		\midrule
		 Gazefollow~\cite{recasens}          & 0.792           & 0.213           & $27.9^{\circ}$             & 0.407                         \\
		 Lian~\cite{lian2018}                & 0.813           & 0.167           & $19.7^{\circ}$             & 0.441                          \\
		 VideoAttention~\cite{chong}         & 0.842           & 0.145           & $16.9^{\circ}$             & 0.476                           \\
		
		\midrule
		\rowcolor{cyan!50}
		HGTTR                 & \textbf{0.912}           & \textbf{0.121}           & $\mathbf{12.7^{\circ}}$            & \textbf{0.538}                           \\
		 	
	\bottomrule[1pt]
		
	\end{tabular}
	\caption{{\bfseries Performance in practical application}. Specifically, `Ang.' refers to the angle between ground-truth gaze direction and the predicted one. }
	\vspace{-0.2cm}
	\label{tab:prat}
\end{table}
	\begin{table}[t]
	\small
	\setlength{\tabcolsep}{9pt}
	\centering
	\begin{tabular}{lcc}
		\toprule[1pt]
		Methods         & Accuracy$\uparrow$  & $L_2$ Dist.$\downarrow$    \\
		\midrule
		 Random                                        & 50.8           & 286                                \\
		 Fixed bias                                    & 52.4           & 122                                 \\
		 GazeFollow~\cite{recasens}                    & 58.7           & 102                                  \\
		 Gaze+Saliency~\cite{pan2016shallow}           & 59.4           & 83                                    \\
		 Gaze+Saliency+LSTM~\cite{hochreiter1997long}  & 66.2           & 71                                     \\
		 Fan~\cite{fan2018inferring}                   & 71.4           & 62                                      \\
		 Sumer~\cite{sumer2020attention}               & 78.1           & 63                                       \\
		 VideoAttention~\cite{chong}                   & 83.3           & 57                                        \\
		
		\midrule
		\rowcolor{cyan!50}
		HGTTR                 & \textbf{90.4}           & \textbf{46}                               \\
		 	
	\bottomrule[1pt]
		
	\end{tabular}
	\caption{{\bfseries Shared attention detection results on the VideoCoAtt dataset}. The interval detection is evaluated in terms of prediction accuracy while the location task is measured with $L_2$ distance.}
	\vspace{-0.5cm}
	\label{tab:share}
\end{table}
	
	\subsection{Discussion}
	It is our belief that defining HGT detection task as simultaneously detecting human head locations and their gaze targets is more reasonable, since manually labeled head locations are unavailable in practical applications. Moreover, joint detection could benefit both of these two tasks with the essential contextual information modeling. However, as a typical Transformer-based method, HGTTR suffers from slow convergence, since it takes long training epochs to learn attention weights to focus on sparse meaningful locations. A possible solution to this issue is designing a more flexible self-attention mechanism with more considerations on this task, like deformable DETR~\cite{deformableDETR}. We will devote more studies on this in the future work.
	
	\noindent
	{\bfseries Broader impacts.} The proposed method predicts human's gaze target. As a human-centric task, it may has some issues about privacy protection when being applied practically, which warrants more policy reviews when using this work in real world applications.
	
	\vspace{-0.2cm}
	\section{Conclusion}
	We have presented an new Transformer-based method for the task of gaze following detection. Our model is designed to allow detect all people's head positions as well as their gaze targets simultaneously. Without the limitation of taking head image as input, our model achieve higher effectiveness and efficiency. Extensive experiments validate its strong performance as well as the potential for understanding gaze behavior in naturalistic human interactions. We hope our method will be useful for human activity understanding research.
	
	\noindent
	{\bfseries Acknowledgments.} This work was supported by NSFC 61831015, National Key R\&D Program of China 2021YFE0206700, NSFC 62176159, Natural Science Foundation of Shanghai 21ZR1432200 and Shanghai Municipal Science and Technology Major Project 2021SHZDZX0102.
	\newpage
	{\small
		\bibliographystyle{ieee_fullname}
		\bibliography{egbib}
	}
	
\end{document}